\pdfoutput=1

\documentclass[11pt]{article}

\usepackage[final]{acl}

\usepackage{times}
\usepackage{latexsym}
\usepackage{enumitem}
\usepackage{booktabs}
\usepackage{multirow}
\usepackage{cleveref}
\crefformat{section}{\S#2#1#3} \crefformat{subsection}{\S#2#1#3}
\crefformat{subsubsection}{\S#2#1#3}
\usepackage{pifont}\usepackage{xcolor}
\definecolor{blue}{RGB}{66,133,244}
\definecolor{red}{RGB}{219,68,55}
\definecolor{green}{RGB}{15,157,88}
\usepackage{algorithm}\usepackage{algpseudocode}

\usepackage[T1]{fontenc}

\usepackage[utf8]{inputenc}

\usepackage{microtype}

\usepackage{inconsolata}

\usepackage{graphicx}

\title{Action Controlled Paraphrasing}

\author{Ning Shi\qquad Zijun Wu \\
  Alberta Machine Intelligence Institute (Amii) \\
  Department of Computing Science \\
  University of Alberta, Edmonton, Canada \\
  \texttt{\{ning.shi,zijun4\}@ualberta.ca} \\}

\begin{document}
\maketitle

\begin{abstract}
Recent studies have demonstrated the potential to control paraphrase generation, 
such as through syntax, 
which has broad applications in various downstream tasks. 
However, 
these methods often require detailed parse trees or syntactic exemplars, 
countering human-like paraphrasing behavior in language use. 
Furthermore, 
an inference gap exists, 
as control specifications are only available during training but not during inference. 
In this work, 
we propose a new setup for controlled paraphrase generation. 
Specifically, 
we represent user intent as action tokens, 
embedding and concatenating them with text embeddings, 
thus flowing together into a self-attention encoder for representation fusion. 
To address the inference gap, 
we introduce an optional action token as a placeholder that encourages the model to determine the appropriate action independently when users' intended actions are not provided. 
Experimental results show that our method successfully enables precise action-controlled paraphrasing 
and preserves or even enhances performance compared to conventional uncontrolled methods when actions are not given.
Our findings promote the concept of action-controlled paraphrasing for a more user-centered design.
\end{abstract}

\vspace{-1mm}
\section{Introduction}
\vspace{-1mm}

Paraphrase generation, the task of producing texts that convey consistent information but with different wording or structures, has been a longstanding problem in natural language processing (NLP) \citep{mckeown-1983-paraphrasing,meteer-shaked-1988-strategies,barzilay-lee-2003-learning,quirk-etal-2004-monolingual,hassan-etal-2007-unt}. It continues to be an active and appealing research topic in recent years \citep{zhou-bhat-2021-paraphrase, hosking-etal-2022-hierarchical} and plays a crucial role in various applications such as sentence simplification \citep{zhao-etal-2018-integrating}, question answering \citep{dong-etal-2017-learning, gan-ng-2019-improving}, adversarial example generation \citep{iyyer-etal-2018-adversarial}, data augmentation \citep{wei2018fast, kumar-etal-2019-submodular, okur-etal-2022-data}, summarization \citep{peng-etal-2019-text}, dialogue generation \citep{gao-etal-2020-paraphrase}, machine translation \citep{thompson-post-2020-automatic}, and semantic parsing \citep{cao-etal-2020-unsupervised-dual, wu-etal-2021-paraphrasing}.

\begin{figure}[t!]
    \centering
    \includegraphics[width=\linewidth]{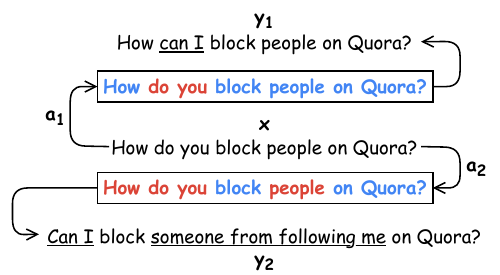}
    \vspace{-6mm}
    \caption{An example from Quora Question Pair dataset showing the high-level overview of action controlled paraphrasing. 
    Given the same text $\mathbf{x}$, the goal is to generate different paraphrase $\mathbf{y}_{1}$ and $\mathbf{y}_{2}$ conditioned on the corresponding actions $\mathbf{a}_{1}$ and $\mathbf{a}_{2}$. 
    It is supposed to keep words in \textcolor{blue}{\textbf{blue}} but to paraphrase those in \textcolor{red}{\textbf{red}}.}
    \label{figure:general}
    \end{figure}

Advanced neural approaches have significantly contributed to this field \citep{NIPS2017_7181,devlin-etal-2019-bert}, 
where 
the sequence-to-sequence framework \citep{sutskever2014sequence} has achieved impressive results \citep{prakash-etal-2016-neural,Wang_Gupta_Chang_Baldridge_2019, witteveen-andrews-2019-paraphrasing, niu-etal-2021-unsupervised}. 
Some have explored other methodologies, 
including variational autoencoders \citep{NIPS2014_d523773c, Gupta_Agarwal_Singh_Rai_2018, bao-etal-2019-generating, roy-grangier-2019-unsupervised}, generative adversarial networks \citep{NIPS2014_5ca3e9b1, yang-etal-2019-end, vizcarra-ochoa-luna-2020-paraphrase}, 
reinforcement learning \citep{li-etal-2018-paraphrase, 10.1145/3394486.3403231}, 
and information retrieval \citep{kazemnejad-etal-2020-paraphrase}. 

Recent studies have gradually shifted to incorporate user control \citep{iyyer-etal-2018-adversarial,chen-etal-2019-controllable,goyal-durrett-2020-neural,kumar-etal-2020-syntax,hosking-etal-2022-hierarchical}. 
These works aim to generate paraphrases that satisfy specific control requirements such as a syntax template \citep{iyyer-etal-2018-adversarial} or a sentential exemplar \citep{chen-etal-2019-controllable}. 
However, these methods often lack user-friendliness. 
They require users to provide detailed control specifications, 
even when such specifications are unnecessary. 
They are counter-intuitive to natural human behavior in paraphrase generation. 
Additionally, an inference gap exists. 
It is often assumed that control specifications are always available, 
which is typically true only during training but not inference or real-world application. 
These limitations highlight the need for more user-friendly and flexible approaches to controlled paraphrase generation.

In this work, we propose action-controlled paraphrasing as a new setup to address these challenges. 
Specifically, we represent users' intentions as action tokens, indicating words to keep or paraphrase.
As additional supervision signals, 
they can be automatically derived from the differences between source texts and target paraphrases without external toolkits or manual labor. 
We introduce action embeddings alongside word and positional embeddings, 
concatenating all three and feeding them into a self-attention encoder for representation fusion.
To close the inference gap, 
we employ a special action token for the model to independently determine the appropriate actions without explicit guidance from the user.
Experimental results demonstrate the effectiveness of our methods. 
On one hand, the specific action control allows users to impose constraints and promote diverse paraphrasing. 
On the other hand, the optional action control balances the trade-off between the control and performance.
Our approach also holds potential for applications such as lexical substitution \citep{qiang2023parals} and adversarial example generation \citep{iyyer-etal-2018-adversarial}, 
which targets the paraphrasing of specific fragments of text.

Overall, our contributions are as follows:\footnote{
Our code and data are publicly available on GitHub: 
\href{https://github.com/ShiningLab/Action-Controlled-Paraphrasing}{github.com/ShiningLab/Action-Controlled-Paraphrasing}.
} 
\begin{enumerate}[leftmargin=*,noitemsep,topsep=0pt]
    \item We propose the action-controlled paraphrasing, a new and more user-friendly setting.
    \item We introduce an end-to-end pipeline, as well as inference alignment, to close the inference gap.
    \item We evaluate both specific and optional action control on common benchmarks, demonstrating their effectiveness.
    \item We further conduct extensive analysis and examine how the actions guide models in the generation of paraphrases.
\end{enumerate}

\section{Related Work}

\textbf{Controlled paraphrase generation} has always been a challenging topic \citep{10.1145/3617680}. 
Diverse paraphrases are believed to be more beneficial to various downstream applications \citep{mccarthy2009components,dong-etal-2017-learning,iyyer-etal-2018-adversarial,NEURIPS2018_398475c8,zhao-etal-2018-integrating,Park_Hwang_Chen_Choo_Ha_Kim_Yim_2019,peng-etal-2019-text}, which motivates many studies to investigate the control over, for example, the syntax. Syntax knowledge often comes from the encoding of either a syntactic form \citep{iyyer-etal-2018-adversarial,huang-chang-2021-generating,hosking-etal-2022-hierarchical} or an exemplar \citep{NIPS2017_2d2c8394,chen-etal-2019-controllable,kumar-etal-2020-syntax,peng-etal-2019-text}. Since inputs contain both the source texts and the syntax specifications, training controlled paraphrase networks usually requires a large amount of annotated data \citep{iyyer-etal-2018-adversarial,chen-etal-2019-controllable} and labor work for annotation \citep{chen-etal-2019-controllable}. The adoption of external tools (e.g., a back translator or a syntax parser) may bring the noise to the data and then mislead the model \citep{wubben-etal-2010-paraphrase,iyyer-etal-2018-adversarial,huang-chang-2021-generating}. Although the control signals can come from more aspects apart from syntax \citep{NIPS2017_2d2c8394,8736871,goyal-durrett-2020-neural,bandel-etal-2022-quality}, from the users' perspective, the need to provide very detailed information is not user-friendly.
\\
\textbf{Inference gap} often exists when the control signals are easily obtained from the training samples but not available for inference. Taking syntactically controlled paraphrasing as an example \citep{iyyer-etal-2018-adversarial}, we can employ an external parser \citep{manning-etal-2014-stanford} during training. However, when it comes to inference, it is found that the trained networks struggle to behave normally when the full parses are replaced with high-level parse templates. A potential solution is to train another model so as to complete full parse trees \citep{iyyer-etal-2018-adversarial,huang-chang-2021-generating}. Nevertheless, the introduction of more models may also bring more noise, and, at least, a template is still required from the user end. A similar problem exists in the control over an exemplar \citep{chen-etal-2019-controllable,kumar-etal-2020-syntax} or a word reordering \citep{goyal-durrett-2020-neural} as well. Another way to address the gap is to relax the control from, for example, a particular parse tree to either keywords \citep{8736871} or a quality vector \citep{bandel-etal-2022-quality}. However, from a user-centered point of view, it comes up with a question can these constraints be optional?

\begin{figure*}[t!]
    \centering
    \includegraphics[width=\linewidth]{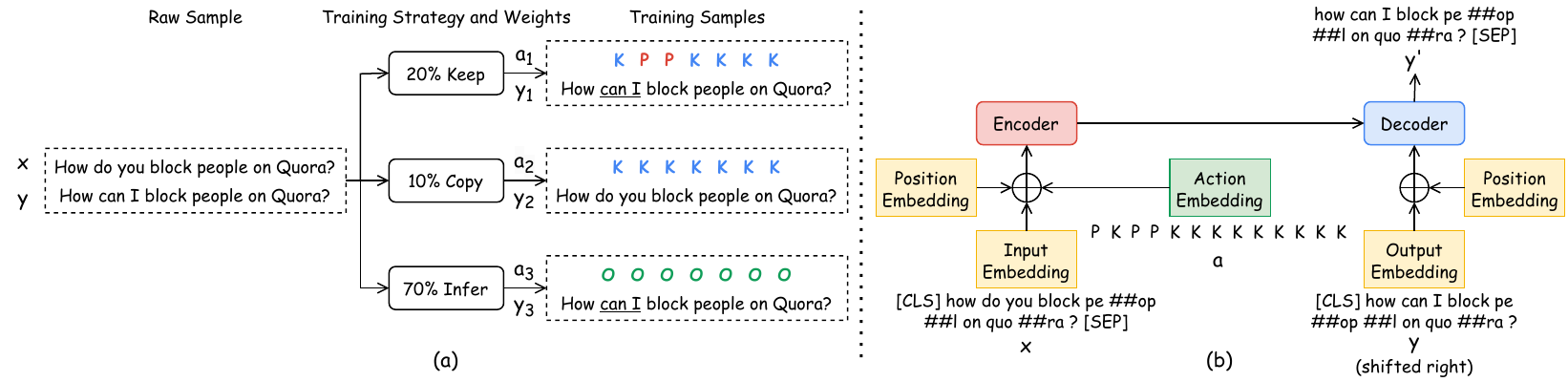}
    \vspace{-7mm}
    \caption{
    (a) explains the proposed training strategy with specific weights. Each action is represented as a sequence in colors, where $K$ means to keep, red $P$ means to paraphrase, and green $O$ denotes a special action $O$ for inference alignment;
    (b) illustrates action embedding in the training pipeline. The encoder input is the concatenation of three states, that is, the input embedding of $\mathbf{x}$, the corresponding positional encoding, and the action embedding of $\mathbf{z}$. 
        In practice, ``{[CLS]}'' is marked as $P$ to remove and ``{[SEP]}'' is marked as $K$ to reserve.}
    \label{figure:method}
    \vspace{-4mm}
\end{figure*}

\vspace{-1mm}
\section{Action Control}
\vspace{-1mm}

We are motivated to offer users a more friendly and intuitive method for controlling the process. Actions serve as a more accessible specification for control compared to other alternatives, such as specific parse templates or exemplars. Users can effortlessly assign desired actions to words, indicating their intended actions and guiding the model to focus on specific parts for paraphrasing. In this section, we provide a detailed description of our specific and optional action control.

\vspace{-1mm}
\subsection{Specific action control}

Paraphrase generation is a task $\mathbf{x} \mapsto \mathbf{y}$ that involves mapping an input sequence $\mathbf{x} \in \mathcal{X}$ to an output sequence $\mathbf{y} \in \mathcal{Y}$, aiming to preserve the semantic content while altering the wording. Here, we introduce a token-level action sequence $\mathbf{a} \in \mathcal{A}$, where $\mathbf{a}=a_{i\leq m}$ and $a\in\{K, P\}$. Each action $\mathbf{a}$ corresponds to a token in the source sequence $\mathbf{x}=x_{i\leq m}$, where $m$ denotes the length of both sequences. Tokens $x_{i}$ are marked with either $K$ or $P$, signifying the intention to keep ($K$) or paraphrase ($P$) them. By that, we denote the paraphrase generation with specific action control as $(\mathbf{x}, \mathbf{a}) \mapsto \mathbf{y}$, where a model is trained to generate the paraphrase $\mathbf{y}$ by considering both the input sequence $\mathbf{x}$ and the action sequence $\mathbf{a}$. An illustrative example is provided in Figure~\ref{figure:general}, where the demonstration is presented at the word level for clarity.

We would like to emphasize that our action control mechanism operates as a soft constraint, requiring models to interpret and incorporate the provided action information into the paraphrasing process.
This means that the model's behavior is influenced by the action guidance, but it retains some flexibility. When the provided action is not reasonable or conflicts with preserving critical information, such as attempting to remove important named entities, the model may prioritize the retention of such words, even if they are assigned the action $P$ to be removed from the output paraphrases. In these cases, the model exercises its discretion to maintain coherence and convey the necessary meaning. The soft constraint nature strikes a balance between adhering to the provided action guidance while considering the overall coherence and fidelity of the generated paraphrases.

\vspace{-1mm}
\subsection{Action generation}

Existing syntactically controlled paraphrasers \citep{iyyer-etal-2018-adversarial,kumar-etal-2020-syntax,huang-chang-2021-generating} often rely on external parsers \citep{manning-etal-2014-stanford}, making them sensitive to the quality of the parse predictor. In contrast, our approach does not require any domain knowledge, external toolkits, or manual labor. The action $\mathbf{a}$ is automatically derived from the input sequence $\mathbf{x}$ and target sequence $\mathbf{y}$, without the need for external assistance.

We start by collecting the vocabulary sets $\mathbf{v}_{\mathbf{x}}$ for the source sequence $\mathbf{x}$ and $\mathbf{v}_{\mathbf{y}}$ for the target sequence $\mathbf{y}$. The intersection of these sets yields the token set $\mathbf{v}_{\mathbf{x}\mathbf{y}}=\mathbf{v}_{\mathbf{x}}\cap\mathbf{v}_{\mathbf{y}}$, which contains all tokens shared by $\mathbf{x}$ and $\mathbf{y}$, including special tokens like the beginning-of-sentence (e.g., "[CLS]") and end-of-sentence (e.g., "[SEP]"). We assign the action $K$ to each token $x_{i}$ in $\mathbf{v}_{\mathbf{x}\mathbf{y}}$, while the remaining tokens are assigned the action $P$. This process results in the action sequence $\mathbf{a}$ that is ensured to correctly satisfy $(\mathbf{x}, \mathbf{a}) \mapsto \mathbf{y}$. The fully automatic nature allows us to generate the action sequence on the fly. To provide an example of a processed action sequence, we present Figure~\ref{figure:method} (b), illustrating how the action sequence is applied. For more details, please refer to Algorithm~\ref{alg:action} in the Appendix.

\vspace{-1mm}
\subsection{Action injection}
\vspace{-1mm}

Following the Transformer architecture \citep{NIPS2017_7181}, we introduce action embeddings alongside word and positional embeddings. Action embeddings serve as a plug-in representation that introduces control signals to the model.

Specifically, within the three embedding layers preceding the encoder, we employ the input embedding and position embedding to process the input sequence $\mathbf{x}$, generating the input encoding and positional encoding, respectively. We then utilize the action embedding layer to produce the action encoding for the action sequence $\mathbf{a}$. To fuse these encoding states, we concatenate them along the sequential dimension. This fusion step combines the input encoding, positional encoding, and action encoding into a unified representation. The resulting fused representation serves as the final input for the encoder, capturing both the semantic content of the input sequence and the action control signals.

One advantage of the action embedding is its straightforward adaptability to various encoder-decoder models, including pre-trained language models. Figure~\ref{figure:method} (b) provides a visual illustration of the architecture. While we focus on simplicity in this work, exploring more advanced fusion structures remains an avenue for future research.

\vspace{-1mm}
\subsection{Optional action control}
\vspace{-1mm}

To address the mismatch between training and inference scenarios, we propose an inference alignment technique that allows for optional action control. We introduce an additional action token, denoted as $O$, representing the optional action. Tokens for which users are uncertain about the appropriate action are tagged with $O$. By assigning $O$ as the action for these tokens, we empower the model to independently determine the eventual action.

For models to understand the proposed optional action control, we adopt a training strategy similar to that used in language model training \citep{devlin-etal-2019-bert}, where we do not always provide the label action. Instead, we employ an action generator that behaves as follows for a certain portion of training instances:
\begin{itemize}[leftmargin=*,noitemsep,topsep=0pt]
    \item \textbf{Keep-20\%}: we retain the label action sequence $\mathbf{a}$ without any modification.
    \item \textbf{Copy-10\%}: we replace the action sequence $\mathbf{a}$ with a copy one, where all action tokens are assigned $K$. Simultaneously, we consider the input sequence $\mathbf{x}$ as the target sequence $\mathbf{y}$, allowing the model to learn this specific copy behavior.
    \item \textbf{Inference-70\%}: We transform the action sequence $\mathbf{a}$ into an optional one, where all action tokens are assigned $O$.
\end{itemize}
Unlike $K$ and $P$, the $O$ action serves a specific purpose in inference alignment. When the target paraphrase is unknown, we can utilize an action sequence consisting entirely of $O$ action tokens. This allows us to compose inputs without explicit action guidance from the user's end.

It is important to highlight that we can leverage an action sequence comprised solely of $K$ to control the model to retain the source words as much as possible. Conversely, an action sequence consisting entirely of $P$ encourages the trained model to generate more diverse paraphrases. Figure~\ref{figure:method} (a) provides an example illustrating this concept.

By employing the $O$ action for inference alignment and utilizing action sequences of all $K$ or all $P$, we enable greater control over the model's behavior during paraphrase generation, allowing for a balance between faithful retention of the source words and the generation of diverse paraphrases.

\vspace{-1mm}
\section{Experiments}
\vspace{-1mm}

The section below describes the evaluation of our method under specific or optional action control.

\begin{table*}[t]
  \begin{center}
  \resizebox{\linewidth}{!}{
  \begin{tabular}{llccccccccc}
    \toprule
    \textbf{Data} & \textbf{Mask} & \textbf{iBLEU-0.8} & \textbf{iBLEU-0.9} & \textbf{BLEU-2} & \textbf{BLEU-3} & \textbf{BLEU-4} & \textbf{METEOR} & \textbf{ROUGE-1} & \textbf{ROUGE-2} & \textbf{ROUGE-L} \\
    \midrule
    \multirow{5}{*}{\texttt{Quora\textsubscript{1}}} & Random & $13.50_{\pm0.42}$ & $19.36_{\pm0.45}$ & $47.66_{\pm0.56}$ & $34.57_{\pm0.53}$ & $25.23_{\pm0.49}$ & $43.35_{\pm0.64}$ & $45.84_{\pm0.65}$ & $22.79_{\pm0.55}$ & $43.31_{\pm0.60}$ \\
     & Ks & $7.50_{\pm0.04}$ & $21.97_{\pm0.04}$ & $61.66_{\pm0.07}$ & $47.39_{\pm0.07}$ & $36.44_{\pm0.07}$ & $60.53_{\pm0.06}$ & $62.39_{\pm0.06}$ & $37.69_{\pm0.06}$ & $59.27_{\pm0.06}$ \\
     & Ps & $9.98_{\pm0.58}$ & $13.04_{\pm0.76}$ & $34.60_{\pm1.23}$ & $23.51_{\pm1.09}$ & $16.11_{\pm0.94}$ & $25.78_{\pm1.77}$ & $26.13_{\pm1.84}$ & $10.41_{\pm1.16}$ & $24.42_{\pm1.75}$ \\
     & Os & $18.49_{\pm0.18}$ & $26.49_{\pm0.28}$ & $58.52_{\pm0.50}$ & $44.79_{\pm0.49}$ & $34.49_{\pm0.47}$ & $56.11_{\pm0.51}$ & $58.74_{\pm0.53}$ & $34.78_{\pm0.54}$ & $56.00_{\pm0.51}$ \\
     & Oracle & $\mathbf{31.11_{\pm0.32}}$ & $\mathbf{40.49_{\pm0.45}}$ & $\mathbf{72.50_{\pm0.54}}$ & $\mathbf{59.71_{\pm0.57}}$ & $\mathbf{49.86_{\pm0.58}}$ & $\mathbf{67.81_{\pm0.45}}$ & $\mathbf{70.90_{\pm0.49}}$ & $\mathbf{51.17_{\pm0.62}}$ & $\mathbf{68.34_{\pm0.50}}$ \\
    \midrule
    \multirow{5}{*}{\texttt{Quora\textsubscript{2}}} & Random & $11.34_{\pm0.39}$ & $16.41_{\pm0.56}$ & $43.14_{\pm0.84}$ & $30.40_{\pm0.80}$ & $21.47_{\pm0.73}$ & $37.66_{\pm0.96}$ & $40.41_{\pm0.87}$ & $18.23_{\pm0.82}$ & $38.18_{\pm0.84}$ \\
     & Ks & $11.83_{\pm0.03}$ & $26.39_{\pm0.05}$ & $66.42_{\pm0.08}$ & $52.12_{\pm0.09}$ & $40.96_{\pm0.09}$ & $64.92_{\pm0.08}$ & $66.62_{\pm0.08}$ & $42.81_{\pm0.09}$ & $63.52_{\pm0.08}$ \\
     & Ps & $7.82_{\pm0.57}$ & $10.32_{\pm0.78}$ & $29.84_{\pm1.45}$ & $19.51_{\pm1.21}$ & $12.81_{\pm0.99}$ & $19.43_{\pm2.13}$ & $19.17_{\pm2.47}$ & $6.07_{\pm1.29}$ & $17.95_{\pm2.32}$ \\
     & Os & $15.91_{\pm0.11}$ & $24.02_{\pm0.17}$ & $55.68_{\pm0.42}$ & $42.20_{\pm0.39}$ & $32.12_{\pm0.33}$ & $52.50_{\pm0.52}$ & $55.66_{\pm0.52}$ & $32.22_{\pm0.54}$ & $53.25_{\pm0.55}$ \\
     & Oracle & $\mathbf{27.15_{\pm0.27}}$ & $\mathbf{36.44_{\pm0.36}}$ & $\mathbf{68.69_{\pm0.48}}$ & $\mathbf{55.73_{\pm0.48}}$ & $\mathbf{45.73_{\pm0.45}}$ & $\mathbf{64.24_{\pm0.50}}$ & $\mathbf{67.80_{\pm0.52}}$ & $\mathbf{47.56_{\pm0.57}}$ & $\mathbf{65.39_{\pm0.52}}$ \\
    \midrule
    \multirow{5}{*}{\texttt{Twitter}} & Random & $21.56_{\pm0.50}$ & $28.91_{\pm0.75}$ & $53.58_{\pm0.87}$ & $43.43_{\pm0.97}$ & $36.26_{\pm1.03}$ & $45.61_{\pm0.78}$ & $50.90_{\pm0.90}$ & $36.42_{\pm1.17}$ & $48.24_{\pm0.97}$ \\
     & Ks & $3.98_{\pm0.14}$ & $17.92_{\pm0.06}$ & $51.45_{\pm0.06}$ & $40.07_{\pm0.05}$ & $31.86_{\pm0.04}$ & $45.85_{\pm0.07}$ & $51.82_{\pm0.07}$ & $33.72_{\pm0.05}$ & $48.39_{\pm0.05}$ \\
     & Ps & $20.10_{\pm0.48}$ & $26.18_{\pm0.63}$ & $49.02_{\pm0.81}$ & $39.18_{\pm0.79}$ & $32.26_{\pm0.80}$ & $41.07_{\pm0.88}$ & $46.38_{\pm0.91}$ & $32.22_{\pm0.95}$ & $43.78_{\pm0.87}$ \\
     & Os & $25.76_{\pm0.29}$ & $34.15_{\pm0.44}$ & $59.52_{\pm0.42}$ & $49.57_{\pm0.54}$ & $42.54_{\pm0.63}$ & $51.04_{\pm0.36}$ & $55.89_{\pm0.62}$ & $42.47_{\pm0.80}$ & $53.43_{\pm0.66}$ \\
     & Oracle  & $\mathbf{34.31_{\pm1.66}}$ & $\mathbf{43.58_{\pm1.75}}$ & $\mathbf{68.53_{\pm1.72}}$ & $\mathbf{59.30_{\pm1.76}}$ & $\mathbf{52.84_{\pm1.84}}$ & $\mathbf{58.33_{\pm1.32}}$ & $\mathbf{63.02_{\pm0.83}}$ & $\mathbf{51.87_{\pm1.25}}$ & $\mathbf{60.83_{\pm0.89}}$ \\
    \bottomrule
  \end{tabular}
  }
  \end{center}
  \vspace{-3mm}
  \caption{The evaluation results for specific action control in \cref{section:exp2}. The model evaluated remains the same TFM*, while the input action used for inference is changed in five ways: Random, Ks, Ps, Os, and Oracle.}
  \label{table:exp1}
  \vspace{-4mm}
\end{table*}

\vspace{-1mm}
\subsection{Data}
\vspace{-1mm}

The following describes two widely-used datasets and how we split and preprocess them into three benchmarks. The data statistics can be found in Table~\ref{table:data}. Please refer to Appendix~\ref{appendix:data} for more details.
\\
\textbf{Quora Question Pair} (QQP) consists of potentially duplicate question pairs, thus being suitable for paraphrase generation.\footnote{\url{kaggle.com/c/quora-question-pairs}} As noted, there is an overlap of source texts between subsets of the standard data split \citep{li-etal-2018-paraphrase}. Therefore, we process the raw dataset in two ways. In \texttt{Quora\textsubscript{1}}, we strictly refer to previous works \citep{li-etal-2019-decomposable,gu2022cdnpg} to locate our implementation better. In \texttt{Quora\textsubscript{2}}, we ensure there is no overlap of source texts among the three subsets, and they remain the same data size. Intuitively, \texttt{Quora\textsubscript{2}} is more challenging than \texttt{Quora\textsubscript{1}} after solving the overlap issue. 
\\
\textbf{Twitter URL} (\texttt{Twitter}) contains tweets sharing the same URLs \citep{lan-etal-2017-continuously}. In line with the others \citep{li-etal-2018-paraphrase,kazemnejad-etal-2020-paraphrase,gu2022cdnpg}, we use the dataset annotated by humans for validation and testing but regard the one labeled automatically as the training set. Hence, \texttt{Twitter} is considered much noisier.

\vspace{-1mm}
\subsection{Setup}
\vspace{-1mm}

Here, we describe the evaluation protocol, model structures, and training configurations.
\\
\textbf{Evaluation.} To better locate ourselves in the big picture, we report evaluation metrics, including BLEU \citep{papineni-etal-2002-bleu}, METEOR \citep{denkowski-lavie-2014-meteor}, ROUGE \citep{lin-2004-rouge}, and iBLEU \citep{sun-zhou-2012-joint}. We refer readers to cited works for an exhaustive explanation. Among all metrics, we consider iBLEU as the primary one. The reason can be found in Appendix~\ref{appendix:eval}. Reported numbers are the results of five runs using random seeds from $[0, 1, 2, 3, 4]$.
\\
\textbf{Model.} We adopt the Transformer (TFM) architecture \citep{NIPS2017_7181} as our base model, following prior works \citep{li-etal-2019-decomposable,gu2022cdnpg}, with the same set of hyperparameters. To explore the effectiveness of action embedding when incorporated with the other pre-trained ones \citep{pennington-etal-2014-glove}, we build a variant TFM\textsubscript{EM} by initializing the embedding layers with pre-trained weights \citep{devlin-etal-2019-bert,wolf-etal-2020-transformers}.\footnote{\url{https://huggingface.co/bert-base-uncased}} Also, we provide two more baselines. In Copy, the model will return the source texts directly. In LSTM, we construct a bi-directional recurrent network \citep{schuster1997bidirectional,hochreiter1997long} with long short-term memory units and an attention mechanism \citep{bahdanau2014neural}. An asterisk (e.g., TFM*) indicates that action control is enabled. For detailed hyperparameters, please refer to Appendix~\ref{appendix:model}.
\begin{table*}[t]
  \begin{center}
  \resizebox{\linewidth}{!}{
  \begin{tabular}{llccccccc}
    \toprule
    \textbf{Data} & \textbf{Model} & \textbf{iBLEU-0.8} & \textbf{BLEU-2} & \textbf{BLEU-4} & \textbf{METEOR} & \textbf{ROUGE-1} & \textbf{ROUGE-2} & \textbf{ROUGE-L} \\
    \midrule
    \multirow{10}{*}{\texttt{Quora\textsubscript{1}}} & TFM \citep{li-etal-2019-decomposable} & $16.25$ & $-$ & $21.73$ & $-$ & $60.25$ & $33.45$ & $-$ \\
     & TFM \citep{zhou-bhat-2021-paraphrase} & $-$ & $42.91$ & $30.38$ & $-$ & $61.25$ & $36.07$ & $-$ \\
     & TFM \citep{gu2022cdnpg} & $21.14$ & $37.97$ & $26.88$ & $38.21$ & $-$ & $-$ & $40.14$ \\
    \cmidrule(r){2-2}
    \cmidrule(r){3-9}
     & Copy & $7.32$ & $62.24$ & $37.05$ & $61.15$ & $63.07$ & $38.40$ & $59.89$ \\
    \cmidrule(r){2-2}
    \cmidrule(r){3-9}
     & LSTM & $15.58_{\pm0.08}$ & $49.69_{\pm0.14}$ & $27.20_{\pm0.16}$ & $44.91_{\pm0.21}$ & $48.89_{\pm0.23}$ & $26.89_{\pm0.20}$ & $47.18_{\pm0.24}$ \\
     & LSTM* & $\mathbf{15.76_{\pm0.10}}$ & $\mathbf{50.97_{\pm0.17}}$ & $\mathbf{28.15_{\pm0.18}}$ & $\mathbf{46.51_{\pm0.27}}$ & $\mathbf{50.71_{\pm0.37}}$ & $\mathbf{28.16_{\pm0.28}}$ & $\mathbf{48.90_{\pm0.37}}$ \\
    \cmidrule(r){2-2}
    \cmidrule(r){3-9}
     & TFM & $18.10_{\pm0.08}$ & $\mathbf{59.36_{\pm0.59}}$ & $\mathbf{34.99_{\pm0.62}}$ & $\mathbf{57.33_{\pm0.48}}$ & $\mathbf{59.91_{\pm0.67}}$ & $\mathbf{35.75_{\pm0.77}}$ & $\mathbf{57.08_{\pm0.75}}$ \\
     & TFM* & $\mathbf{18.49_{\pm0.18}}$ & $58.52_{\pm0.50}$ & $34.49_{\pm0.47}$ & $56.11_{\pm0.51}$ & $58.74_{\pm0.53}$ & $34.78_{\pm0.54}$ & $56.00_{\pm0.51}$ \\
    \cmidrule(r){2-2}
    \cmidrule(r){3-9}
     & TFM\textsubscript{EM} & $\mathbf{18.48_{\pm0.29}}$ & $\mathbf{60.75_{\pm0.58}}$ & $\mathbf{36.36_{\pm0.57}}$ & $\mathbf{58.63_{\pm0.49}}$ & $\mathbf{61.42_{\pm0.58}}$ & $\mathbf{37.46_{\pm0.70}}$ & $\mathbf{58.72_{\pm0.61}}$ \\
     & TFM\textsubscript{EM}* & $18.43_{\pm0.09}$ & $58.39_{\pm0.43}$ & $34.40_{\pm0.42}$ & $55.91_{\pm0.44}$ & $58.59_{\pm0.35}$ & $34.69_{\pm0.52}$ & $55.90_{\pm0.38}$ \\
    \midrule 
    \multirow{7}{*}{\texttt{Quora\textsubscript{2}}} & Copy & $11.84$ & $68.11$ & $42.70$ & $66.69$ & $68.43$ & $44.77$ & $65.17$ \\
    \cmidrule(r){2-2}
    \cmidrule(r){3-9}
     & LSTM & $10.78_{\pm0.09}$ & $39.81_{\pm0.19}$ & $18.96_{\pm0.14}$ & $33.12_{\pm0.25}$ & $36.99_{\pm0.30}$ & $16.81_{\pm0.17}$ & $35.79_{\pm0.28}$ \\
     & LSTM* & $\mathbf{11.34_{\pm0.08}}$ & $\mathbf{41.93_{\pm0.24}}$ & $\mathbf{20.43_{\pm0.19}}$ & $\mathbf{35.73_{\pm0.34}}$ & $\mathbf{40.06_{\pm0.43}}$ & $\mathbf{18.80_{\pm0.30}}$ & $\mathbf{38.72_{\pm0.39}}$ \\
    \cmidrule(r){2-2}
    \cmidrule(r){3-9}
     & TFM & $\mathbf{16.24_{\pm0.12}}$ & $\mathbf{59.62_{\pm0.36}}$ & $\mathbf{35.18_{\pm0.35}}$ & $\mathbf{57.33_{\pm0.40}}$ & $\mathbf{60.70_{\pm0.44}}$ & $\mathbf{36.69_{\pm0.47}}$ & $\mathbf{58.12_{\pm0.46}}$ \\
     & TFM* & $15.91_{\pm0.11}$ & $55.68_{\pm0.42}$ & $32.12_{\pm0.33}$ & $52.50_{\pm0.52}$ & $55.66_{\pm0.52}$ & $32.22_{\pm0.54}$ & $53.25_{\pm0.55}$ \\
    \cmidrule(r){2-2}
    \cmidrule(r){3-9}
     & TFM\textsubscript{EM} & $\mathbf{16.47_{\pm0.22}}$ & $\mathbf{59.71_{\pm0.81}}$ & $\mathbf{35.23_{\pm0.75}}$ & $\mathbf{57.41_{\pm0.76}}$ & $\mathbf{60.79_{\pm1.05}}$ & $\mathbf{36.60_{\pm1.11}}$ & $\mathbf{58.18_{\pm1.13}}$ \\
     & TFM\textsubscript{EM}* & $15.84_{\pm0.30}$ & $55.61_{\pm0.57}$ & $32.03_{\pm0.42}$ & $52.37_{\pm0.75}$ & $55.52_{\pm0.72}$ & $32.31_{\pm0.61}$ & $53.22_{\pm0.71}$ \\
    \midrule
    \multirow{7}{*}{\texttt{Twitter}} & Copy & $3.36$ & $51.61$ & $31.93$ & $46.05$ & $52.06$ & $33.84$ & $48.58$ \\
    \cmidrule(r){2-2}
    \cmidrule(r){3-9}
     & LSTM & $11.37_{\pm0.72}$ & $34.07_{\pm1.40}$ & $19.04_{\pm1.20}$ & $25.95_{\pm1.38}$ & $34.47_{\pm1.60}$ & $18.94_{\pm1.44}$ & $32.68_{\pm1.54}$ \\
     & LSTM* & $\mathbf{12.47_{\pm0.57}}$ & $\mathbf{37.75_{\pm1.17}}$ & $\mathbf{21.84_{\pm1.03}}$ & $\mathbf{29.56_{\pm1.16}}$ & $\mathbf{38.03_{\pm1.20}}$ & $\mathbf{22.06_{\pm1.20}}$ & $\mathbf{36.10_{\pm1.13}}$ \\
    \cmidrule(r){2-2}
    \cmidrule(r){3-9}
     & TFM & $\mathbf{25.79_{\pm0.28}}$ & $\mathbf{59.92_{\pm0.21}}$ & $\mathbf{43.10_{\pm0.16}}$ & $\mathbf{51.61_{\pm0.07}}$ & $\mathbf{56.84_{\pm0.12}}$ & $\mathbf{43.50_{\pm0.08}}$ & $\mathbf{54.42_{\pm0.12}}$ \\
     & TFM* & $25.76_{\pm0.29}$ & $59.52_{\pm0.42}$ & $42.54_{\pm0.63}$ & $51.04_{\pm0.36}$ & $55.89_{\pm0.62}$ & $42.47_{\pm0.80}$ & $53.43_{\pm0.66}$ \\
    \cmidrule(r){2-2}
    \cmidrule(r){3-9}
     & TFM\textsubscript{EM} & $25.24_{\pm0.24}$ & $\mathbf{59.34_{\pm0.53}}$ & $\mathbf{42.34_{\pm0.71}}$ & $\mathbf{51.09_{\pm0.59}}$ & $\mathbf{56.07_{\pm1.05}}$ & $\mathbf{42.62_{\pm1.09}}$ & $\mathbf{53.59_{\pm1.11}}$ \\
     & TFM\textsubscript{EM}* & $\mathbf{25.48_{\pm0.11}}$ & $59.07_{\pm0.29}$ & $42.15_{\pm0.43}$ & $50.72_{\pm0.29}$ & $55.58_{\pm0.49}$ & $42.16_{\pm0.62}$ & $53.10_{\pm0.55}$ \\
    \bottomrule
  \end{tabular}
  }
  \end{center}
  \vspace{-2mm}
  \caption{The evaluation results under optional action control in \cref{section:exp1}. The method Copy stands for copying the inputs directly as the outputs. The first half (1-3 rows) of \texttt{Quora\textsubscript{1}} are borrowed from previous works. Since the evaluation is not consistent, we report as more metrics as we can. We leave it as ``$-$" if a metric is not reported previously. It is noted that the datasets split to train TFM \citep{zhou-bhat-2021-paraphrase} is not clarified, so it might be different. Bold numbers emphasize the higher from with or without optional action control given the same model.}
  \label{table:exp2}
  \vspace{-4mm}
\end{table*}
\\
\textbf{Training.} We train on a 32GB V100 with a batch size of $64$. The AdamW optimizer \citep{loshchilov2018decoupled} is applied with a learning rate of $5e^{-5}$ and an $\ell_2$ gradient clipping of $1.0$ \citep{10.5555/3042817.3043083}. A linear scheduler helps manage the training process after a warm-up of $5000$ steps. We adopt early stopping to wait for a lower validation loss until there are no updates for $32$ epochs \citep{prechelt1998early}. Meanwhile, we keep $256$ as the maximum epochs for training to save time. We believe a more careful tuning of the training configurations can further boost the performance. Yet, given the purpose of validating the assumption of action control, we leave this to future work.

\vspace{-1mm}
\subsection{Experiment: specific action control} \label{section:exp1}
\vspace{-1mm}

\textbf{Setup.} 
The main objective of this experiment is to evaluate the effectiveness of specific action control, with a particular focus on the TFM* model. We aim to validate the specific action control by manipulating the input action, as illustrated in Table~\ref{table:exp1}. We consider the following variations for the input actions in our experiment:
\begin{itemize}[leftmargin=*,noitemsep,topsep=0pt]
    \item Random: actions of random $K$ and $P$.
    \item Ks: actions of only $K$.
    \item Ps: actions of only $P$.
    \item Os: actions of only $O$.
    \item Oracle: gold label actions.
\end{itemize}
By exploring these different input action settings, we can assess the impact of specific action control on the TFM* model's performance.
\\
\textbf{Results.} As expected, Ks forces the model to copy from the source sentence, while Ps encourages the model to revise as many words as possible, resulting in increased diversity. 
Consequently, Ps achieves an iBLEU-0.8 of approximately 9.98 on \texttt{Quora\textsubscript{1}}, up to 2.48 higher than that of Ks. 
However, the BLEU-4 increases from 16.11 to 36.44 when using Ks rather than Ps. 
The disallowance of borrowing source words in Ps leads to the lowest BLEU-4 of 12.81 on \texttt{Quora\textsubscript{2}}, and an iBLEU of 7.82.
In short, Ps often improves the iBLEU, while Ks benefits the BLEU-based metrics, highlighting the effectiveness of specific action control. Furthermore, the overall improvement of Os across all three datasets demonstrates the effectiveness of our proposed inference alignment for optional control. 
Even without detailed guidance, Os still shows significant advancements. The dominance of Oracle over the others is surprising, achieving 31.11, 27.15, and 34.31 in terms of iBLEU-0.8 on \texttt{Quora\textsubscript{1}}, \texttt{Quora\textsubscript{2}}, and \texttt{Twitter} separately. The label actions not only controls the paraphrasing but also guides the model to generate paraphrases that closely match those in the test set. This suggests that the incorporation of search algorithms or similar techniques may further enhance the performance of Os, approaching that of Oracle, which could be interesting for future research.

\vspace{-1mm}
\subsection{Experiment: optional action control} \label{section:exp2}
\vspace{-1mm}

\textbf{Setup.} 
Next, we proceed to explore optional action control in a fair comparison with uncontrolled models. In this scenario, the action for every source text is consistently set to the optional action (i.e., actions consisting only of $O$). We present the experimental results in Table~\ref{table:exp2}.
\\
\textbf{Baselines.} To ensure the correctness of our implementation, we compare the results of our TFM model with those reported in previous works \citep{li-etal-2019-decomposable, zhou-bhat-2021-paraphrase, gu2022cdnpg}. Using a consistent evaluation protocol, we observe that, for iBLEU-0.8, our TFM achieves $18.10$, surpassing the score of $16.25$ reported in \citep{li-etal-2019-decomposable} but falling below $21.14$ reported in \citep{gu2022cdnpg}. For ROUGE-1 and ROUGE-2, our TFM aligns well with that of the others. For BLEU score, our TFM outperforms the others across the board on \texttt{Quora\textsubscript{1}}, achieving a BLEU-2 of 59.36 and a BLEU-4 of 34.99. Such discrepancy could be attributed to factors such as the smoothing function \citep{chen-cherry-2014-systematic} and the wordpiece tokenizer \citep{gu2022cdnpg}. We would like to highlight that a straightforward copy approach is sufficient to achieve satisfactory results, particularly in terms of metrics such as BLEU-4. The Copy baseline achieves the highest BLEU-4 of 37.05 and 42.70, but the lowest iBLEU-0.8 of 7.32 and 11.84 on \texttt{Quora\textsubscript{1}} and \texttt{Quora\textsubscript{2}}, respectively.
\\
\textbf{Results.} A noteworthy finding is the competitive performance observed when enabling optional action control. 
Even without explicit action guidance, TFM* achieves an iBLEU-0.8 score of 18.49 on \texttt{Quora\textsubscript{1}}, surpassing the 18.10 achieved by TFM. Furthermore, on the more challenging \texttt{Quora\textsubscript{2}} and noisier \texttt{Twitter}, the drop in iBLEU-0.8 compared to TFM is minimal, with only a decrease of 0.33 (from 16.24 to 15.91) and 0.03 (from 25.79 to 25.76), respectively.
We notice that the gap between TFM* and the other metrics is more substantial than iBLEU-0.8. Upon closer examination of the generated paraphrases, we confirm that models without control are more prone to copying the source texts, resulting in higher scores for metrics such as BLEU-4 but lower iBLEU-0.8. For example, on \texttt{Quora\textsubscript{2}}, the BLEU-4 for TFM is 35.18, which is 3.06 higher than that of TFM*, yet the iBLEU-0.8 is only 0.33 higher at 16.24. This suggests that various actions during training promote diverse generation and discourage excessive copying of the source texts. Overall, these observations underlines the effectiveness of our inference alignment in enabling optional action control when action specifications are not provided. These findings consistently hold when evaluating LSTM* and TFM\textsubscript{EM}* against LSTM and TFM\textsubscript{EM}. Notably, LSTM* exhibits the most counterintuitive improvement, outperforming LSTM across all three benchmarks in terms of all metrics. We hypothesize that the presence of the action embedding layer enhances the model capacity and contributes to this improvement. However, this enhancement does not appear to occur in more powerful models such as TFM* and TFM\textsubscript{EM}*.

\begin{table}[t]
  \begin{center}
  \resizebox{\linewidth}{!}{
  \begin{tabular}{cl}
    \toprule
    \textbf{Raw} $\mathbf{x}$ & How can I improve my English communication? \\
    \textbf{Raw} $\mathbf{y}$ & How can I increase my English fluency? \\
    \cmidrule(r){1-1}
    \cmidrule(r){2-2}
    $\mathbf{x}$ & {[CLS]} how can i imp \#\#rov my english com \#\#mun \#\#ic ? {[SEP]} \\
    $\mathbf{y}$ & how can i \underline{inc \#\#rea \#\#s} my english \underline{flu \#\#en \#\#ci} ? {[SEP]} \\
    \cmidrule(r){1-1}
    \cmidrule(r){2-2}
    $\mathbf{x_{1}}$ & \textcolor{red}{{[CLS]} how} \textcolor{blue}{can i imp \#\#rov my english com \#\#mun \#\#ic ? {[SEP]}} \\
    $\mathbf{a_{1}}$ & \textcolor{red}{P P} \textcolor{blue}{K K K K K K K K K K K} \\
    $\mathbf{y^{\prime}_{1}}$ & \underline{what} can i \underline{do to} imp \#\#rov my english \underline{speak} ? {[SEP]} \\
    \cmidrule(r){1-1}
    \cmidrule(r){2-2}
    $\mathbf{x_{2}}$ & \textcolor{red}{{[CLS]} how can i imp \#\#rov my english com \#\#mun \#\#ic ?} \textcolor{blue}{{[SEP]}} \\
    $\mathbf{a_{2}}$ & \textcolor{red}{P P P P P P P P P P P P} \textcolor{blue}{K} \\
    $\mathbf{y^{\prime}_{2}}$ & \underline{what are some way to} imp \#\#rov english ? [SEP] \\
    \cmidrule(r){1-1}
    \cmidrule(r){2-2}
    $\mathbf{x_{3}}$ & \textcolor{red}{{[CLS]}} \textcolor{green}{how can i imp \#\#rov my english com \#\#mun \#\#ic ?} \textcolor{blue}{{[SEP]}} \\
    $\mathbf{a_{3}}$ & \textcolor{red}{P} \textcolor{green}{O O O O O O O O O O O} \textcolor{blue}{K} \\
    $\mathbf{y^{\prime}_{3}}$ & how can i imp \#\#rov my com \#\#mun \#\#ic \underline{skill and} english \underline{prof \#\#ici} ? {[SEP]} \\
    \midrule
    \textbf{Raw} $\mathbf{x}$ & What are the plans for the new year? \\
    $\mathbf{x}$ & {[CLS]} what are the plan for the new year ? {[SEP]} \\
    \cmidrule(r){1-1}
    \cmidrule(r){2-2}
    $\mathbf{x_{1}}$ & \textcolor{red}{{[CLS]}} \textcolor{blue}{what are} \textcolor{red}{the} \textcolor{blue}{plan} \textcolor{red}{for the} \textcolor{blue}{new year ? {[SEP]}} \\
    $\mathbf{a_{1}}$ & \textcolor{red}{P} \textcolor{blue}{K K} \textcolor{red}{P} \textcolor{blue}{K} \textcolor{red}{P P} \textcolor{blue}{K K K K} \\
    $\mathbf{y^{\prime}_{1}}$ & {[CLS]} what are \underline{some} plan for \underline{this} new year ? {[SEP]} \\
    \cmidrule(r){1-1}
    \cmidrule(r){2-2}
    $\mathbf{x_{2}}$ & \textcolor{red}{{[CLS]}} \textcolor{blue}{what} \textcolor{red}{are the plan for} \textcolor{blue}{the new year ? {[SEP]}} \\
    $\mathbf{a_{2}}$ & \textcolor{red}{P} \textcolor{blue}{K} \textcolor{red}{ P P P P} \textcolor{blue}{K K K K K} \\
    $\mathbf{y^{\prime}_{2}}$ & {[CLS]} what \underline{is your} new year \underline{res \#\#ol \#\#ut} ? {[SEP]} \\
    \cmidrule(r){1-1}
    \cmidrule(r){2-2}
    $\mathbf{x_{3}}$ & \textcolor{red}{{[CLS]}} \textcolor{green}{what are the plan for the new year ?} \textcolor{blue}{{[SEP]}} \\
    $\mathbf{a_{3}}$ & \textcolor{red}{P} \textcolor{green}{O O O O O O O O O} \textcolor{blue}{K} \\
    $\mathbf{y^{\prime}_{3}}$ & {[CLS]} what \underline{is your} plan for the new year ? {[SEP]} \\
    \bottomrule
  \end{tabular}
  }
  \end{center}
  \vspace{-3mm}
  \caption{Two examples for the case study in \cref{section:case}. Each $\mathbf{y^{\prime}}$ is generated by TFM* given a pair of ($\mathbf{x}$, $\mathbf{a}$). The first example is borrowed from test set of \texttt{Quora\textsubscript{1}}. The second one is manually created by ourselves.}
  \label{table:case}
  \end{table}

\vspace{-1mm}
\section{Analysis}
\vspace{-1mm}

This section comprises a case study, a sensitivity analysis, and an ablation study, all aimed at further investigating the aspects of optional and specific action control in detail.

\vspace{-1mm}
\subsection{Case study} \label{section:case}
\vspace{-1mm}

\begin{figure}[t!]
    \centering
    \includegraphics[width=\linewidth]{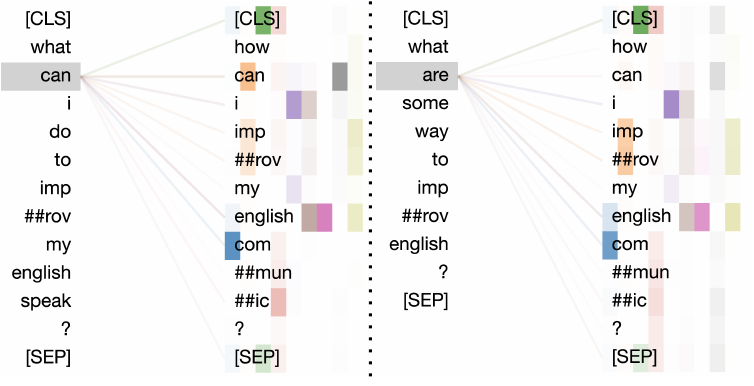}
    \vspace{-8mm}
    \caption{The cross-attention dependencies of the second decoder layer when generating the token ``are'' and ``can'' for paraphrases $\mathbf{y^{\prime}_{1}}$ (left) and $\mathbf{y^{\prime}_{2}}$ (right) of the first case in Table~\ref{table:case}. The colors of each column represent the distribution of an attention head.}
    \label{figure:attention}
    \end{figure}

Table~\ref{table:case} presents two case studies, with the first half selected from the test set of \texttt{Quora\textsubscript{1}}, and the second half manually created by ourselves. These cases serve to illustrate the impact of modifying the input action on the paraphrasing outcomes.

In the first case, we observe clear changes resulting from modifications in the input action. For instance, the change from ``how can i'' to ``what can i do to'' in $\mathbf{y^{\prime}_{1}}$ is the consequence of applying the $P$ action to the token ``how'' in $\mathbf{x_{1}}$. Similarly, in $\mathbf{y^{\prime}_{2}}$, the action $P$ assigned to both ``how'' and ``can'' in $\mathbf{x_{2}}$ leads to the revision of the phrase as ``what are some way to''. Notably, even though the crucial token ``english'' in $\mathbf{x_{2}}$ is marked as $P$, the model still references it in $\mathbf{y^{\prime}_{2}}$ to preserve its semantic meaning. This observation suggests that the action control operates as a soft constraint, allowing some flexibility in adhering to the specified actions.

In the second case, we again observe similar controlled effectiveness. Additionally, we would like to highlight that since we created this case manually, it provides some evidence of the generalizability of action control. This generalization holds implications not only for paraphrase generation but also for the broader application and effectiveness of the control mechanism. Exploring such generalization in future research would be intriguing.

Furthermore, we examine the distribution of cross-attention in the decoder of TFM* when generating the first case. Figure~\ref{figure:attention} illustrates the attention dependencies on the same input sequence $\mathbf{x}$ when generating the words ``can'' and ``are'' in the corresponding paraphrases $\mathbf{y^{\prime}_{1}}$ and $\mathbf{y^{\prime}_{2}}$, respectively. Upon assigning the $P$ action to ``can'', the attention heads that originally focused on ``can'' are partly redirected towards other tokens such as ``imp'', ``\#\#rov'', and ``english''. This shift in attention suggests that the model needs to infer the new word ``are'' based on the context, rather than copying ``can'' from the source sequence. The visualization of the attention distribution provides insights into how the model adapts its attention mechanism to incorporate the specified actions and generate appropriate paraphrases.

\begin{figure}[t!]
    \centering
    \includegraphics[width=\linewidth]{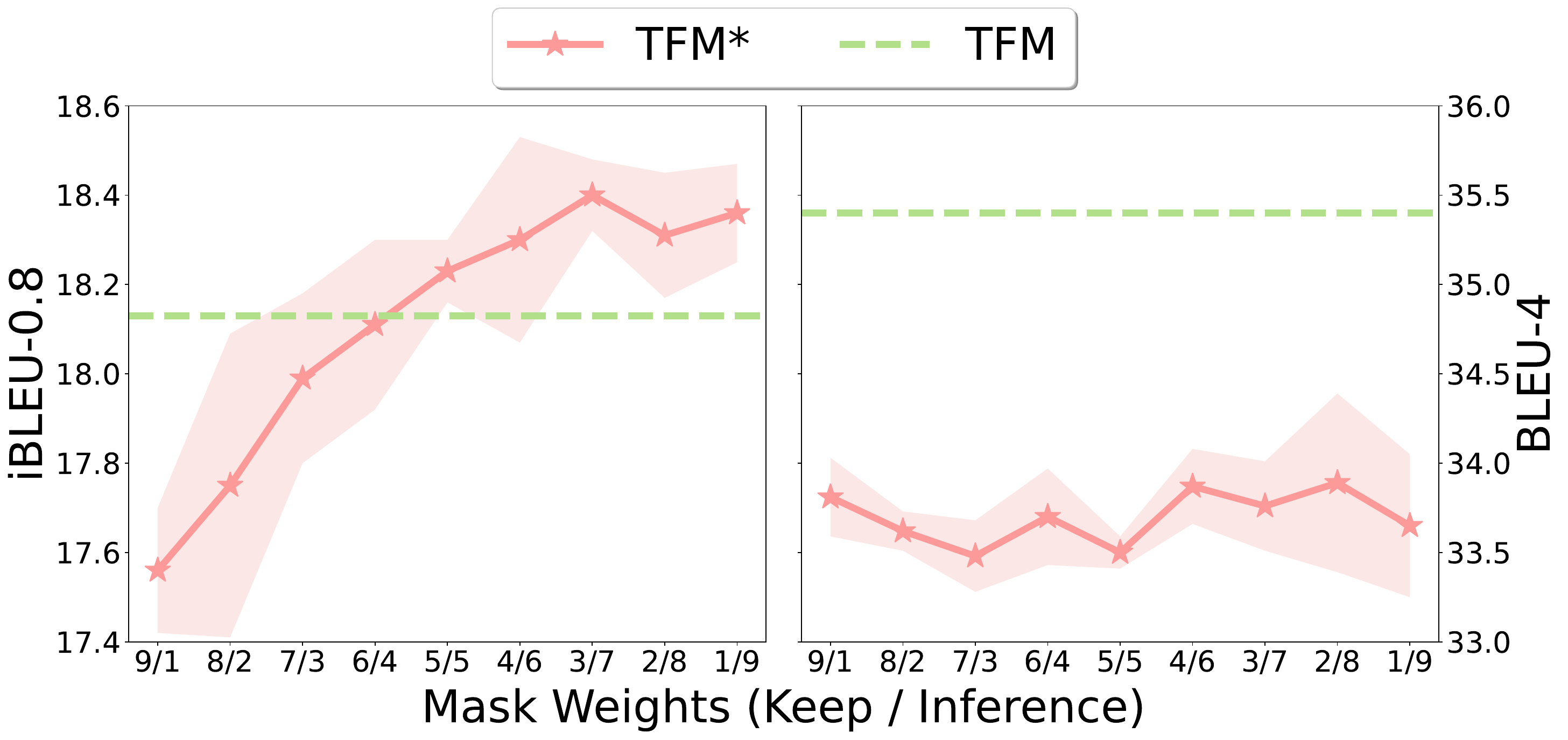}
    \vspace{-7mm}
    \caption{Evaluation of TFM* on \texttt{Quora\textsubscript{1}} with the optional action control. Two metrics, iBLEU-0.8 (left) and BLEU-4 (right), are reported against different action weights standing for the proportion of keeping the true action or replacing it with an optional action. The green dashed line is the baseline TFM.}
    \label{figure:action_weights}
    \end{figure}

\vspace{-1mm}
\subsection{Training weights}
\vspace{-1mm}

The sensitivity analysis allows us to carefully tune the action weights for optional action control. In this analysis, we exclude the Copy operation from the training strategy and adjust the weights for Keep and Inference. By varying the weights from 90/10\% to 10/90\%, we observe the trends in terms of iBLEU-0.8 and BLEU-4, as shown in Figure~\ref{figure:action_weights}.

In the left sub-figure, which focuses on iBLEU-0.8, we observe that TFM* outperforms the baseline TFM when simultaneously decreasing the rate of Keep to 50\% and increasing the rate of Inference to 50\%. This suggests a balanced combination leads to improved performance. In the right sub-figure concerning BLEU-4, the improvement is not as significant with a growing standard deviation. One possible explanation is the incorporation of action control prevents the model from overfitting and encourages the generation of paraphrases that differ from the target paraphrases in the test set. Finally, we determine that the optimal weights for achieving desirable performance are around 30/70\% for Keep and Inference, respectively.

\vspace{-1mm}
\subsection{Training strategy}
\vspace{-1mm}

To explicitly highlight the contribution of the Copy operation in the training strategy, we perform an ablation study. Here, we remove the Copy operation and redistribute its 10\% weight to either Keep or Inference. The model used here is the same TFM*, and the evaluation is conducted on \texttt{Quora\textsubscript{1}} using optional action control.
Surprisingly, the results (partially shown in Table~\ref{table:strategy}) indicate all metrics benefit from Copy. This finding supports our initial assumption that the Copy operation teaches the model to reuse words when encountering the $K$ action, particularly in comparison to samples where the input text remains the same but the action sequence is filled with $P$. This also highlights the bias inherent in evaluating the quality of paraphrases solely based on BLEU-related scores, as copying source words often leads to better outcomes.

\begin{table}[t]
  \begin{center}
  \resizebox{\linewidth}{!}{
  \begin{tabular}{ccccccc}
    \toprule
    \textbf{Keep} & \textbf{Copy} & \textbf{Inference} & \textbf{iBLEU-0.8} & \textbf{BLEU-4} & \textbf{METEOR} & \textbf{ROUGE-L} \\
    \midrule
    30\% & 0\% & 70\% & $18.40_{\pm0.08}$ & $33.76_{\pm0.25}$ & $54.95_{\pm0.22}$ & $55.06_{\pm0.21}$ \\
    20\% & 10\% & 70\% & $\mathbf{18.49_{\pm0.18}}$ & $\mathbf{34.49_{\pm0.47}}$ & $\mathbf{56.11_{\pm0.51}}$ & $\mathbf{56.00_{\pm0.51}}$ \\
    20\% & 0\% & 80\% & $18.31_{\pm0.14}$ & $33.89_{\pm0.50}$ & $55.18_{\pm0.56}$ & $55.25_{\pm0.59}$ \\
    \bottomrule
  \end{tabular}
  }
  \end{center}
  \vspace{-3mm}
  \caption{Comparison of TFM* on \texttt{Quora\textsubscript{1}} with and without Copy operation in the training strategy. The evaluation is under the optional action control. The second line using 20\%, 10\%, and 70\% is the reported version in the main experiments.}
  \label{table:strategy}
  \end{table}

\vspace{-2mm}
\section{Conclusion}
\vspace{-2mm}

Overall, 
we have presented action-controlled paraphrasing, 
allowing users to selectively determine which words to keep or paraphrase. 
We propose an end-to-end procedure for representing user intent as actions 
and seamlessly incorporating them into the input encoding through representation fusion. 
During training, 
the action generation process is fully automated, 
requiring no domain knowledge, external toolkits, or manual labor. 
To mirror human-like adaptability, 
we incorporate a special action token into the training strategy, 
enabling models to make independent decisions without user guidance 
and thus bridging the gap between training and inference. 
Empirical evaluations demonstrate the effectiveness of our approach under both specific and optional action control. 
Specific action control achieves precise paraphrasing aligned with input actions, 
and optional action control balances the trade-off between the control and performance. 
Our framework promotes the adoption of optional control in future research, 
aligning more closely with human cognitive processes, 
and enhancing user experiences in paraphrase generation.

\section*{Limitations}
The action control framework proposed in this work primarily guides paraphrase generation based on the words selected by users from the source sentence. Consequently, our approach may not be ideally suited for handling cases where the desired paraphrase is longer than the source sentence. In such scenarios, the target paraphrase could potentially include all the words from the source sentence, resulting in all words being tagged with the keep action. This situation could potentially mislead the model into returning the input sentence as the output paraphrase, without any meaningful transformation. By tackling this challenge, we can enhance the applicability and effectiveness of our approach in handling a wider range of paraphrase generation scenarios.

\section*{Acknowledgment}
We gratefully appreciate Bradley Hauer, Grzegorz Kondrak, and Lili Mou for sharing their pearls of wisdom.
This work is based on a course project for CMPUT651: Deep Learning for NLP at the University of Alberta.

\bibliography{custom}

\vfill
\pagebreak

\appendix
\clearpage

\section{Related Work}

While our work shares some similarities with the study conducted by \citet{NEURIPS2019_5e2b6675}, there are several key distinctions that set our research apart. Firstly, the primary objectives of our respective studies differ significantly. Our method focuses on a new control setting, rather than a pure generation model. As such, it is not a matter of one approach being superior to the other. Secondly, during training, we derive the action based on not the the latent bag of words but the differences between the source and target sentences. Lastly, our research emphasizes the importance of determining which source words to retain or revise, even without access to the target information. This focus on source words is a key aspect of our method that differentiates it from the previous one. In short, while both studies contribute valuable insights to the field of paraphrase generation, our work introduces a unique approach that emphasizes user control and the importance of an action-oriented generation process.

\section{Data} \label{appendix:data}

\textbf{Quora Question Pair} (QQP) consists of 404,290 lines of potentially duplicate question pairs. Among them, 149,263 pairs are annotated as paraphrases of each other, thus being suitable as a benchmark for paraphrase generation. As noted \citep{li-etal-2018-paraphrase}, there is an overlap of source texts between subsets of the standard data split. Therefore, we split the dataset using the following two ways for a fair comparison. In \texttt{Quora\textsubscript{1}}, we are strictly in line with previous works \citep{li-etal-2019-decomposable,gu2022cdnpg}, dividing the dataset into three subsets for training, validation, and testing after a random shuffle. In \texttt{Quora\textsubscript{2}}, we first group all the sample pairs with one-to-one mapping, yielding 62,544 in total. By that, we separate 4K as the validation set and 20K as the test one. Then, the remaining 38,544 samples join the many-to-many group to compose the training pool containing 12,3626 samples. From the pool, we randomly select 100K as the eventual training set. By that, we ensure there is no overlap of source texts among the three subsets, and they remain the same data size. Intuitively, \texttt{Quora\textsubscript{2}} is more challenging than \texttt{Quora\textsubscript{1}} since models can not see the source text in the test set during training.
\\
\textbf{Twitter URL} (\texttt{Twitter}) is a large-scale sentential paraphrase collected from Twitter by connecting tweets through shared URLs \citep{lan-etal-2017-continuously}. It provides two subsets, where one is annotated by humans, and the other one is labeled automatically. Following the other works \citep{li-etal-2018-paraphrase,kazemnejad-etal-2020-paraphrase,gu2022cdnpg}, we randomly sample 1K and 5K from the former as the validation and test sets and 110K from the latter as the training set. Hence, Twitter is considered noisier because the training labels are generated by algorithms. We find overlap among subsets as well. However, we can not conduct a new split since, after excluding duplicates, the rest samples are insufficient to maintain the same training size.
\\
\textbf{Preprocessing} is applied to all datasets unless specific notes. Words are stemmed and converted into lowercase \citep{li-etal-2019-decomposable}. A normalizer clears texts by unifying accented characters (e.g., from Café to cafe) and replacing spaces with classic ones. Similar to others \citep{li-etal-2018-paraphrase, li-etal-2019-decomposable, kazemnejad-etal-2020-paraphrase, gu2022cdnpg}, we truncate all the tokenized sentences (sequences) longer than 20.

\begin{table}[t]
\centering
\resizebox{\linewidth}{!}{
  \begin{tabular}{lcccccc}
    \toprule
    \multirow{2}{*}{\textbf{Data}} & \multicolumn{3}{c}{\textbf{Subset}} & \multicolumn{3}{c}{\textbf{Seq. Len.}} \\
    \cmidrule(r){2-4}
    \cmidrule(r){5-7}
     & \textbf{\#Train.} & \textbf{\#Valid.} & \textbf{\#Test} & \textbf{Max.} & \textbf{Min.} & \textbf{Avg.} \\
    \cmidrule(r){1-1}
    \cmidrule(r){2-4}
    \cmidrule(r){5-7}
    \texttt{Quora\textsubscript{1\&2}} & 100K & 4K & 20K & 20 & 2 & 12.80 \\
    \texttt{Twitter} & 110K & 1K & 5K & 20 & 3 & 17.46 \\
    \bottomrule
  \end{tabular}}
    \caption{Data statistics of \texttt{Quora\textsubscript{1}}, \texttt{Quora\textsubscript{2}}, and  \texttt{Twitter} regarding data size and sequence length.}
  \label{table:data}
\end{table}

\begin{table*}[t]
\centering
\resizebox{\linewidth}{!}{
  \begin{tabular}{lcccccccccc}
    \toprule
    \multirow{2}{*}{\textbf{Model}} & \multirow{2}{*}{\textbf{Embedding}} & \multicolumn{4}{c}{\textbf{Encoder}} & \multicolumn{4}{c}{\textbf{Decoder}} & \multirow{2}{*}{\textbf{Parameters}} \\
    \cmidrule(r){3-6}
    \cmidrule(r){7-10}
     & & Layer & Hidden & Attention Head & Feedforward & Layer & Hidden & Attention Head & Feedforward \\
    \midrule
    LSTM & 450 & 3 & 450 & - & - & 1 & 450 & - & - & 41.23M \\
    TFM & 450 & 3 & 450 & 9 & 1024 & 3 & 450 & 9 & 1024 & 48.75M \\
    TFM\textsubscript{EM} & 768 & 3 & 768 & 12 & 1024 & 3 & 450 & 9 & 1024 & 58.45M \\
    \bottomrule
  \end{tabular}}
    \caption{Hyperparameters of models LSTM, TFM, and TFM\textsubscript{EM}.}
  \label{table:model}
\end{table*}

\section{Evaluation} \label{appendix:eval}

To better locate ourselves in the big picture, we report evaluation metrics as more as possible, including BLEU \citep{papineni-etal-2002-bleu}, METEOR \citep{denkowski-lavie-2014-meteor}, ROUGE \citep{lin-2004-rouge}, and iBLEU \citep{sun-zhou-2012-joint}. The n-gram to calculate the BLEU score will be specified when being reported. Among all metrics, we consider iBLEU as the primary one. In paraphrase generation, usually, $\mathbf{x}$ and $\mathbf{y}$ are partly the same. A simple copy of $\mathbf{x}$ as the prediction $\mathbf{y^{\prime}}$ can produce a satisfactory performance in terms of conventional evaluation metrics (e.g., BLEU). Instead, iBLEU penalizes the similarity between $\mathbf{y^{\prime}}$ and $\mathbf{x}$. The $\alpha$ to calculate the iBLEU score will be specified in the report. We refer readers to the related works mentioned above for an exhaustive explanation. All the reported numbers attached with standard deviation are based on the results of five runs using random seeds from $[0, 1, 2, 3, 4]$.

\begin{algorithm}[htp!]
    \small
    \caption{Action Generation}
    \label{alg:action}
    \renewcommand{\algorithmicrequire}{\textbf{Input:}}
    \renewcommand{\algorithmicensure}{\textbf{Output:}}
    \begin{algorithmic}[1]
        \Require
            Source texts $\mathbf{X}$ and target paraphrases $\mathbf{Y}$.
        \Ensure
            Actions $\mathbf{A}$.
        \State $\mathbf{A} \gets \emptyset$
        \For{$(\mathbf{x},\mathbf{y}) \in (\mathbf{X}, \mathbf{Y})$}
            \State $\mathbf{a} \gets \emptyset$
            \State ($\mathbf{v}_{\mathbf{x}}, \mathbf{v}_{\mathbf{y}}) \gets [$set$(\mathbf{x})$, set$(\mathbf{y})]$
            \State $\mathbf{v}_{\mathbf{x}\mathbf{y}} \gets \mathbf{v}_{\mathbf{x}}\cap\mathbf{v}_{\mathbf{y}}$
            \For{$x \in \mathbf{x}$}
                \If{$x$ in $\mathbf{v}_{\mathbf{x}\mathbf{y}}$}
                    \State $\mathbf{a} \gets \mathbf{a} \cup [K]$
                \Else
                    \State $\mathbf{a} \gets \mathbf{a} \cup [P]$
                \EndIf
            \EndFor
            \State $\mathbf{A} \gets \mathbf{A} \cup [\mathbf{a}]$
        \EndFor
        \State \Return $\mathbf{A}$
    \end{algorithmic}
\end{algorithm}

\section{Model} \label{appendix:model}

We implement our pipeline and models on the top of the Huggingface \citep{wolf-etal-2020-transformers}. To align with previous works \citep{li-etal-2019-decomposable,gu2022cdnpg}, we focus on the Transformer (TFM) \citep{NIPS2017_7181} with the same hyperparameters. As a tiny version, TFM has $3$ layers for the encoder and decoder, where each layer has $9$ attention heads with a hidden size of $450$. The dimension of the embedding and feedforward layers are $450$ and $1024$. There are a total of $41.23$M trainable parameters. To better understand the efficacy of action embedding when incorporated with the other pre-trained ones \citep{pennington-etal-2014-glove}, we build a variant TFM\textsubscript{EM} by initializing the embedding layers with pre-trained weights from BERT \citep{devlin-etal-2019-bert}. As a result, TFM\textsubscript{EM} has an embedding size of $768$. Its encoder has $12$ attention heads with a hidden size of $768$. The other structure and hyperparameters of TFM\textsubscript{EM} remain the same as TFM. There are $58.45$M trainable parameters in total. Beam search is utilized for both TFM and TFM\textsubscript{EM} during inference with 8 beams. Also, we provide two more baseline models. In Copy, the model will return the source texts as paraphrases by direct copying. In LSTM, we construct a bi-directional recurrent network \citep{schuster1997bidirectional,hochreiter1997long} with long short-term memory units and an attention mechanism \citep{bahdanau2014neural}. Its encoder consists of $3$ layers with a hidden size of $225$ in each direction, and its decoder has one layer with a hidden size of $450$. The embedding size is $450$ for both the encoder and decoder. There are a total of $48.75$M trainable parameters. Teacher forcing with a rate of $0.5$ serves to spur up the training process \citep{williams1989learning}. The vocab size for all models is fixed as $30522$ as we use the same wordpiece tokenizer. An asterisk (i.e., *) after the model name (e.g., TFM*) indicates that action control is enabled.

\end{document}